# Reducing the Computational Cost in Multi-objective Evolutionary Algorithms by Filtering Worthless Individuals

**Zahra Pourbahman[1], Ali Hamzeh[2]**

[1] Department of Electronic and Computer Engineering, Shiraz University,
Shiraz, Iran

[2] Department of Electronic and Computer Engineering, Shiraz University,
Shiraz, Iran

**Abstract**
The large number of exact fitness function evaluations makes evolutionary algorithms to have computational cost (especially in Multi Objective Problems (MOPs)). In some real- world problems, reducing number of these evaluations is much more valuable even by increasing computational complexity and spending more time. To fulfil this target, we introduce an effective factor, in spite of applied factor in Adaptive Fuzzy Fitness Granulation NSGAII (AFFG_NSGAII) algorithm, to filter out worthless individuals more precisely. Our proposed approach is compared with respect to AFFG_NSGAII, using the Hypervolume (HV) and the Inverted Generational Distance (IGD) performance measures. The proposed method is applied to 1 traditional and 1 state-of-the-art benchmarks with considering 3 different dimensions. From an average performance view, the results indicate that although decreasing the number of fitness evaluations leads to have performance reduction but it is not tangible compared to what we gain.

## 1. Introduction

Evolutionary algorithms (EAs) have a population-based procedure in which the population evolves repeatedly by applying some stochastic operators in order to generate better population members until a termination control criterion is met [1]. They seem to be one of promising optimizers among recent proposed optimization methods [2] since they have a number of unique features as follows. (i) EAs can be implemented simply [1], (ii) EAs can find multiple optimal solutions ideally while classical optimization methodologies can't find such solutions efficiently [3], and (iii) EAs perform the parallel search as a computationally quick procedure in contrast with classical optimization methodologies [1].

In a wide variety of real-world optimization problems EAs can be applied [1] because most of the time, practical problems have two or even more normally conflicting objectives which should be optimized simultaneously as MOPs in which a set of optimal solutions (effective solutions) needs to be obtained and EAs can find these effective solutions efficiently in a single run [2] whereas they use a population-based approach. This trend is known as Multi Objective Evolutionary Algorithms (MOEAs). Also, set of these effective solutions is known as Pareto optimal set and their vectors are called non-dominated. Non-dominated vectors are plotted in objective space and constituted the Pareto front. In MOPs, to assign a fitness value to an individual, all objectives should be evaluated. Therefore, when MOEAs are applied to a complex problem, computational cost can grow increasingly [4]. Also, it can be time-consuming. To handle these difficulties, fitness approximation can be integrated into MOEAs [5].

In our work, we aim to reduce the number of exact fitness function evaluations in one of the state-of-the-art proposed approaches for fitness approximation, which is called AFFG_NSGAII [6, 7] by introducing an effective factor to diagnose valuable individuals more logically.

The remainder of this paper proceeds as follows. Literature review is provided in Section 2. Section 3 expresses the contributions of the proposed approach. Section 4 and Section 5 present the experimental setup and the experimental results respectively. Section 6 is devoted to discussion. Finally, Conclusion of the paper and future directions are described in Section 7.

## 2. Related Work

In some real-world problems, metaheuristics like evolutionary algorithms are used to find a set of solutions over a unique run. The large number of exact fitness function evaluations makes such problems computationally prohibitive. The computational cost becomes more critical in MOPs since more objectives are involved. To deal with this difficulty, it is common to use approximation techniques, which are divided into three levels, namely, problem approximation, functional approximation, and evolutionary approximation [8].

Problem approximation tries to substitute an easier computationally solvable problem for the original one. As an example, performance evaluation of turbine blade wind tunnel experiments, which is led to Euler equations, can be replaced with Computational Fluid Dynamics (CFD) simulation, which is led to Navier-Stokes equations [8, 9].

Functional approximation tries to estimate a model based on objective functions known as the fitness function in the evolutionary computation [8]. To approximate fitness function, surrogate-assisted evolutionary computation can be used [9]. In recent researches, an Aggregate Surrogate Model (ASM) for multi objective optimization is introduced [10, 11, 12]. This surrogate model is built based on the combination of One-Class Support Vector Machine (SVMs) to change (randomly) unsupervised population into supervised one in each generation and Support Vector Regression (SVR) to estimate fitness of each individual and provide a Pareto front at last. Since ASM is used for extrapolation, to have enough diversity in the search space, informed operators are applied. In the other hand, because there are some errors in the surrogate model, pre-children generated by informed operators are not filtered in the basis of ASM lonely. In [10, 11] it is done based on ASM gain with regard to the lowest distance to non-dominated solutions in each generation. In [12] offsprings are filtered in a greedy manner (The highest amount of {ASM (k) – ASM ($z_k$) | k is an offspring and $z_k$ is the nearest neighbour of k}), but premature convergence forces to make a probabilistic selection besides by using a normal distribution. Another research on surrogate-assisted MOEAs was proposed in [13]. In this work a Pareto Rank Learning procedure is used to predict rank of each new offspring. To learn this surrogate model, a population evolves iteratively and the value of each new offspring is evaluated using original objective functions and then is archived into an embedded database. Archiving new offsprings continue until having enough training data. After that, the non-dominance sorting is applied over the archived solutions. Then, an ordinal SVM model is learned based on the sorted database. At last, the rank of each new offspring is predicted in terms of the learned model; if the model produces an output of rank 1 its fitness is evaluated and is archived in the database. This updated database is used for updating the model after each generation. Even though fitness function approximation models presented can reduce cost of solving problems with expensive objective functions, they have some general defects as follows. Since model is updated in each generation, computational burden rises. Moreover, the precision of the model is dependant to the primary training data. Additionally, complexity of model grows exponentially as the number of problem parameters increases [14].

Evolutionary approximation specifically is used in evolutionary algorithms. Fitness inheritance is a major class of this type of approximation, which was basically introduced in [15]. In this method, the number of fitness approximation in contrast with fitness evaluation is controlled based on *inheritance proportion* parameter. For an individual, its similarity to its parents and fitness of its parents are applied to form a weighted average formula for fitness approximation. Despite the simplicity of this method, since similarity of each individual to its parents is evaluated just in the decision space, its performance is not acceptable [14].

To address the above-mentioned difficulties (in functional and evolutionary approximation), a new method for fitness approximation based on information granulation was introduced by Davarynejad et al. [7] Called Adaptive Fuzzy Fitness Granulation NSGAII (AFFG_NSGAII). In this method, a pool of solutions is constituted in the objective space. Each member of the pool is called a fuzzy granule. Each fuzzy granule is a Gaussian Membership Function (GMF) where its center is an individual and its width is computed based on its fitness and some problem dependant parameters. But approximated fitness sometimes leads to have not sufficient precision in such calculations. So, the weighted rank of each member of the pool besides a problem dependant parameter which is the minimum width of GMFs is used instead [6]. Additionally, each fuzzy granule has a life index used in fitness approximation while it can control the computational complexity of the algorithm. In this method, fitness of each new individual, generated by an evolutionary algorithm (like the standard NSGAII) in the decision space, being approximated or evaluated explicitly is determined based on its maximum similarity to the granules of the pool. The maximum similarity is evaluated in terms of a predefined similarity metric, which is Gaussian similarity function. In this criterion, the computed width of each fuzzy granule is used as a parameter for controlling the degree of the similarity among a new individual and that fuzzy granule; if the maximum similarity of the new individual to the granules of the pool be lower than an adaptive threshold, its fitness is approximated by increasing the granule's life index. As a point, in [6] the threshold is considered fixed (0.9) to simplify their evaluations. In the other hand, in AFFG_NSGAII, the pool size is controlled in terms of the life index parameter in which the granule with the lowest life index is removed from the pool when its size becomes more than a predefined threshold.

Since AFFG_NSGAII is one of the state-of-the-art proposed approaches in fitness approximation, vast researches over it was done recently. For example, as mentioned earlier, by the reason that the width of each granule is used in the similarity metric as an important factor, in [16] a fuzzy logic controller is embedded to propose a width for each fuzzy granule. Input of this controller is Number of Decision Variable (NDV), Maximum Range of Decision Variables (MRDV), and number of completed generations. As an application-based work, effectiveness of the granule-based fitness approximation on Spread Spectrum Watermarking was investigated in [17].

In some applicatory problems like simulation, reducing number of expensive objective function evaluations is taken into consideration. AFFG_NSGAII can be used to deal with this limitation as a recent and promising method for fitness approximation. However, in this method, exploring valuable solutions for fitness evaluations more accurately can have a considerable effect on reducing the number of exact fitness function evaluations without sacrificing its performance. We contribute to this area and acquire some achievements.

## 3. The Proposed Approach

As mentioned in the previous section, AFFG_NSGAII as an evolutionary algorithm integrated into fitness approximation was improved in [6], in which filtering worthless individuals for fitness approximation was performed in terms of a fixed threshold and an applied factor that is a similarity metric called Gaussian similarity function. It means that the value of each new offspring generated by the standard NSGAII is determined based on its maximum similarity to the granules of the pool which is computed by the similarity metric; if the maximum similarity of a new offspring to the pool be lower than a fixed threshold, it is considered as a valuable individual. So, it is added to the pool and its fitness is evaluated explicitly; otherwise, fitness of the most similar granule of the pool is assigned to the new offspring (fitness is approximated) and then granule's life index is increased.

However, we found that in such evolutionary process, there are some potential factors despite applied factor to filter out worthless individuals more precisely. Even if applying these factors leads to have an increase in the computational complexity but decreasing the computational cost in many applicatory problems, in which there are multiple expensive-to-evaluate objective functions, is much more significant under a limited resource budget. In our proposed approach, we introduce the most effective factor among the potential factors in spite of Gaussian similarity function, to identify more logically whether a new offspring is worthy enough for exactly evaluating its fitness. Before the introduction of the new effective factor, a preprocessing should be explained in the following.

In each generation, granules of the pool are ranked based on non-dominance sorting [18]. Then, Non-inferior solutions are considered as the Current Pareto Set.

Inspired by the fact that in most MOEAs the population is driven toward the best Pareto points [10], we introduce an influential factor in order to guide the search in the vicinity of the Current Pareto Set in each generation.

Suppose the phenotype of *jth* individual and the center of *lth* fuzzy granule in *ith* generation to be like

$$X_j^{(i)} = \{x_{j,1}^{(i)}, x_{j,2}^{(i)}, \ldots, x_{j,n}^{(i)}\},$$

$$C_l^{(i)} = \{c_{l,1}^{(i)}, c_{l,2}^{(i)}, \ldots, c_{l,n}^{(i)}\}$$

, respectively and *d* be considered as the dimension of each individual, the minimum distance between *jth* individual and *k'* elements of the Current Pareto Set in *ith* generation is computed based on Euclidean distance, as Eqn. (1).

$$D_{j,l}\left(X_j^{(i)}, C_l^{(i)}\right) = \sqrt{\sum_{n=1}^{d}\left(X_{j,n}^{(i)} - C_{l,n}^{(i)}\right)^2} \quad (1)$$

, where $l = 1, 2, \ldots, k'$

Now assume that the maximum similarity of the new offspring to the granules of the pool be lower than the determined threshold like before. To decide fitness of that offspring is either evaluated or approximated, the minimum distance of both the new offspring and its parents to the Current Pareto Set are computed by Eqn. (1); if the new offspring be closer to the Current Pareto Set compared with at least one of its parents, the new offspring is considered as a valuable individual, is added to the pool as a new fuzzy granule and its fitness is evaluated explicitly. We call our proposed approach as Modified_AFFG_NSGAII.

Even if applying our powerful factor leads to make algorithm more complicated but in some real-world problems like expensive simulation-based and mechanical design problems decreasing the computational cost is much more considerable even by increasing computational complexity and spending more time.

In Modified_AFFG_NSGAII, we use our influentially promising factor, which is the minimum distance to the Current Pareto Set despite applied factor, thereby deciding that fitness of a new offspring is either evaluated or approximated. Consequently, the estimation of the proximity of solutions to the real Pareto set locally leads to have more precise selections of valuable individuals for fitness evaluations. In this way, we deal with the computational cost burden of such expensive problems by remarkably reducing the number of exact fitness function evaluations without having any tangible effect in the viewpoints of efficiency and efficacy.

In order to prove that our proposed approach is promising, 14 test problems are applied. Additionally, 2 well-known performance metrics are used for validation of our proposed approach.

## 4. The Experimental Setup

This section describes comprehensive assessments by means of two well-known performance metrics and adopting wide varieties of test problems to compare our results with respect to those obtained with a state-of-the-art algorithm for fitness approximation (AFFG_NSGAII) [6].

### 4.1 Performance Measures

In this section, we present 2 indicators, which are commonly used specially in MOEAs for evaluation of our proposed approach.

### 4.1.1 Hypervolume

For a minimization problem, the volume in the objective space covered by non-inferior solutions (N) is evaluated by this metric. The set of the worst values of objectives forms a vector as the reference set. As explained mathematically by Deb in [2], for each non-inferior solution, $s \in N$, a hypercube, $V_s$, is constructed with a reference point, r. After all, Hypervolume is calculated based on the union of all hypercubes, as follow:

$$HV = Volume\left(U_{s=1}^{|N|} V_s\right) \qquad (2)$$

To make it sensible, it is showed in Fig 1 [2].

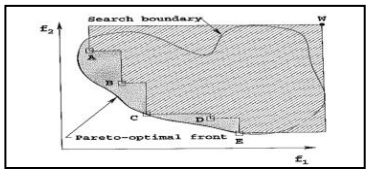

Fig.1  The Hypervolume enclosed by non-dominated solutions.

### 4.1.2 Inverted Generational Distance

A real Pareto front and a set of candidate solutions

$$PF = \{y_1, y_2, \ldots, y_N\},$$
$$F = \{X_1, X_2, \ldots, X_k\}$$

are given; the Inverted Generational Distance (IGD) is defined as follow:

$$IGD(F, PF) = \frac{1}{N}\left(\sum_{j=1}^{N} \acute{D}_j^t\right)^{\frac{1}{t}}, 1 \leq t \leq \infty \qquad (3)$$

Where $\acute{D}_j$ is minimal Euclidean Distance from $y_j$ to F [19].

### 4.2 Benchmarks

In this section, we present 1 traditional and 1 state-of-the-art benchmarks in order to perform comprehensive assessments of our proposed approach.

### 4.2.1 Congress on Evolutionary Computation 2009 (CEC09)

In the CEC 2009 algorithm competition, a set of bound constrained MOP test problems as UF family and a set of constrained test problems as CF family are suggested [20]. In our experiments we adopt 5 test problems from CF family, CF1 to CF5, and 4 test problems from UF family, UF1 to UF5 except UF4.

### 4.2.2 Zitzler-Deb-Thiele (ZDT)

As it was emerged in [21], ZDT family test problems have sufficient complexity to compare different types of multi objective optimizers. In our experiments, we adopt 5 test problems, ZDT1 to ZDT6 except ZDT5 as a binary problem.

## 5. The Experimental Results

Some parameter settings need to be performed in our experiments. The population size is set to 50. A set of new offsprings are generated by Simulated Binary Crossover (SBX) with probability of 0.9 and Polynomial Mutation (PM) with the probability of 1/L, where L is the number of decision variables. Distribution indices for crossover and mutation are taken from the literature ($\eta_c = 20$ and $\eta_m = 20$). Furthermore, binary tournament selection is applied. Tables 1, 2, 3 show amounts of mentioned design parameters per test problem.

Table 1: Utilized parameter values and reference points used for calculating $I_H$ in ZDT family and their number of decision variables.

| Problem | $\sigma_{min}$ | $N_G$ | Reference point | Number of decision variables |
|---|---|---|---|---|
| ZDT1 | $2^{-4}$ | 100 | [1.1,3.5] | 6 |
| ZDT2 | $2^{-5}$ | 100 | [1.1,5.0] | 6 |
| ZDT3 | $2^{-5}$ | 100 | [1.1,6.0] | 6 |
| ZDT4 | $2^{-6}$ | 100 | [1.1,140] | 10 |
| ZDT5 | $2^{-5}$ | 100 | [1.1,9.0] | 10 |

Table 2: Utilized parameter values and reference points used for calculating $I_H$ in CF family and their number of decision variables.

| Problem | $\sigma_{min}$ | $N_G$ | Reference point | Number of decision variables |
|---|---|---|---|---|
| CF1 | $2^{-4}$ | 100 | [3,3] | 10 |
| CF2 | $2^{-4}$ | 100 | [8,7] | 10 |
| CF3 | $2^{-4}$ | 100 | [68,59] | 10 |
| CF4 | $2^{-4}$ | 100 | [18,19] | 10 |
| CF5 | $2^{-4}$ | 100 | [31,32] | 10 |

Table 3: Utilized parameter values and reference points used for calculating $I_H$ in UF family and their number of decision variables.

| Problem | $\sigma_{min}$ | $N_G$ | Reference point | Number of decision variables |
|---|---|---|---|---|
| UF1 | $2^{-4}$ | 100 | [8,7] | 30 |
| UF2 | $2^{-4}$ | 100 | [6,5] | 30 |
| UF3 | $2^{-4}$ | 100 | [12,10] | 30 |
| UF5 | $2^{-4}$ | 100 | [18,15] | 30 |

Additionally, all numerical results are the average of 30 independent runs, which are presented in Tables 4, 5, 6, 7, 8, and 9. These attainments are related to HV and IGD of Pareto front while ZDT, CF, and UF test problem families are applied, respectively and both AFFG_NSGAII and Modified_AFFG_NSGAII converge. Indeed, these results indicate that both methods approximately have the same performance.

Table 4: IGD of both AFFG_NSGAII and Modified_AFFG_NSGAII after convergence in ZDT family.

| Problem | AFFG_NSGAII IGD | Modified_AFFG_NSGAII IGD |
|---|---|---|
| ZDT1 | 0.0364 | 0.0370 |
| ZDT2 | 0.0235 | 0.0254 |
| ZDT3 | 0.0284 | 0.0381 |
| ZDT4 | 2.6393 | 3.2972 |
| ZDT6 | 1.5948 | 1.7387 |

Table 5: HV of both AFFG_NSGAII and Modified_AFFG_NSGAII after convergence in ZDT family.

| Problem | AFFG_NSGAII HV | Modified_AFFG_NSGAII HV |
|---|---|---|
| ZDT1 | 3.4501 | 3.4504 |
| ZDT2 | 4.7906 | 4.7874 |
| ZDT3 | 6.6715 | 6.6600 |
| ZDT4 | 150.4215 | 149.7011 |
| ZDT6 | 5.1082 | 5.1251 |

Table 6: IGD of both AFFG_NSGAII and Modified_AFFG_NSGAII after convergence in CF family.

| Problem | AFFG_NSGAII IGD | Modified_AFFG_NSGAII IGD |
|---|---|---|
| CF1 | 0.6615 | 0.5578 |
| CF2 | 0.2872 | 0.3018 |
| CF3 | 1.0518 | 1.2223 |
| CF4 | 0.2580 | 0.2981 |
| CF5 | 1.4652 | 1.6357 |

Table 7: HV of both AFFG_NSGAII and Modified_AFFG_NSGAII after convergence in CF family.

| Problem | AFFG_NSGAII HV | Modified_AFFG_NSGAII HV |
|---|---|---|
| CF1 | 8.4249 | 8.3584 |
| CF2 | 53.8690 | 53.3946 |
| CF3 | 3.9161e+003 | 3.8987e+003 |
| CF4 | 334.5758 | 332.9370 |
| CF5 | 928.7142 | 929.7592 |

Table 8: IGD of both AFFG_NSGAII and Modified_AFFG_NSGAII after convergence in UF family.

| Problem | AFFG_NSGAII IGD | Modified_AFFG_NSGAII IGD |
|---|---|---|
| UF1 | 0.2389 | 0.3185 |
| UF2 | 0.1134 | 0.1456 |
| UF3 | 0.4946 | 0.6081 |
| UF4 | 2.1404 | 2.2922 |

Table 9: HV of both AFFG_NSGAII and Modified_AFFG_NSGAII after convergence in UF family.

| Problem | AFFG_NSGAII HV | Modified_AFFG_NSGAII HV |
|---|---|---|
| UF1 | 54.4400 | 53.5675 |
| UF2 | 28.8462 | 28.7701 |
| UF3 | 112.5421 | 110.7408 |
| UF4 | 218.7533 | 217.6718 |

In the other hand, the average number of exact fitness function evaluations (of 30 independent runs) was plotted against the number of generations, which is determined in terms of the convergence time per test problem per algorithm. Derivative Figures (from Fig. 2 to Fig. 7) signify that Modified_AFFG_NSGAII reduces the computational cost considerably compared with AFFG_NSGAII.

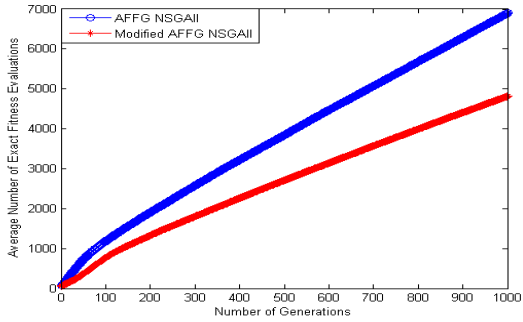

Fig. 2 Computational Cost Comparison of Modified_AFFG_NSGA2 and AFFG_NSGA2 over ZDT1 Problem.

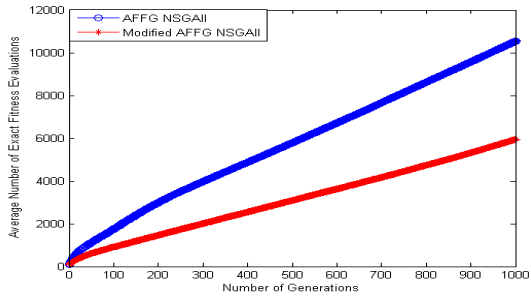

Fig. 3 Computational Cost Comparison of Modified_AFFG_NSGA2 and AFFG_NSGA2 over ZDT4 Problm

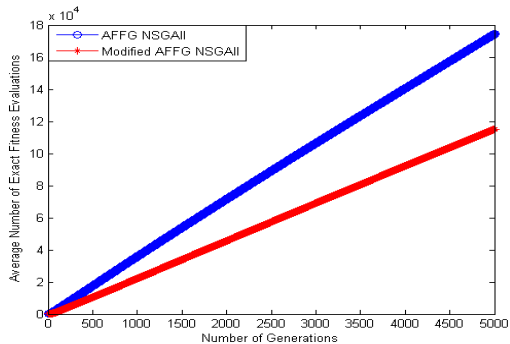

Fig. 4 Computational Cost Comparison of Modified_AFFG_NSGA2 and AFFG_NSGA2 over CF1 Problem.

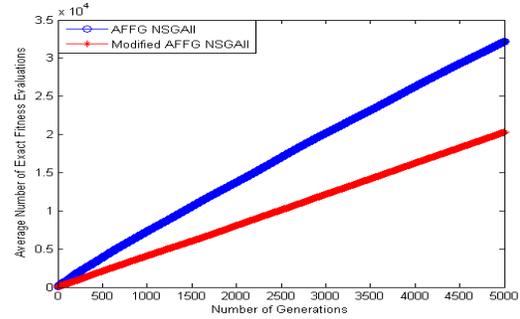

Fig 5. Computational Cost Comparison of Modified_AFFG_NSGA2 and AFFG_NSGA2 over CF3 Problem.

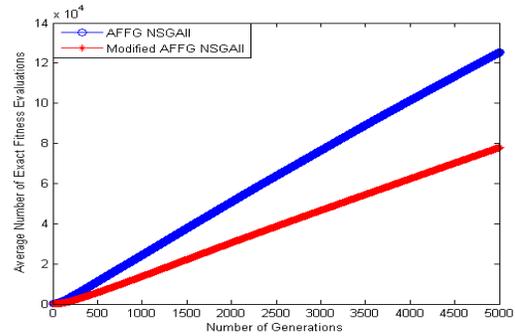

Fig.6 Computational Cost Comparison of Modified_AFFG_NSGA2 and AFFG_NSGA2 over UF1 Problem.

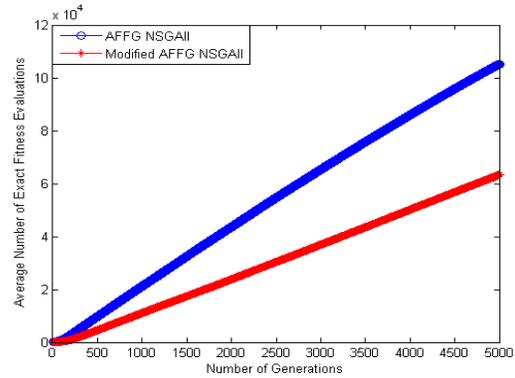

Fig.7 Computational Cost Comparison of Modified_AFFG_NSGA2 and AFFG_NSGA2 over UF3 Problem.

## 6. Discussion

As some state-of-the-art MOEAs integrated with fitness approximation in the literature, it is common to perform fitness approximation for some individuals besides fitness evaluation as usual for others. In some applicatory problems like simulation-based and mechanical design problems, there are expensive objective functions to evaluate. Therefore, contribution to this area has attracted more attention, recently. In this paper, we have contributed to this area in order to decrease the computational cost.

We believe that if we have had even lower individuals for fitness evaluations as usual in each generation but higher confidence about their qualities, termination control criterion would be met sooner. Therefore, there is a trade-off between the computational cost and the computational complexity to achieve this fidelity. To achieve the above target, inspired by the fact that in most MOEAs the population is driven toward the best Pareto points, we proposed an effective and powerful factor in order to guide the search in the vicinity of the Current Pareto Set in each generation. Derivative Figures (Fig. 8 to Fig. 19) prove that Modified_AFFG_NSGAII mostly outperforms AFFG_NSGAII in terms of HV and IGD metrics per adopted test problem.

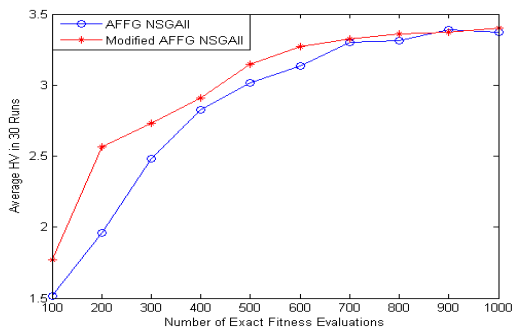

Fig.8  Performance (HV) Comparison of Modified_AFFG_NSGA2

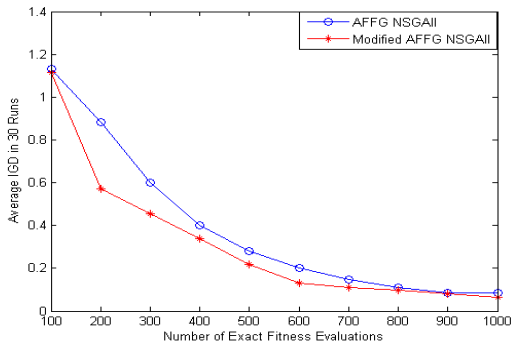

Fig.9  Performance (IGD) Comparison of Modified_AFFG_NSGA2 and AFFG_NSGA2 over ZDT1 Problem.

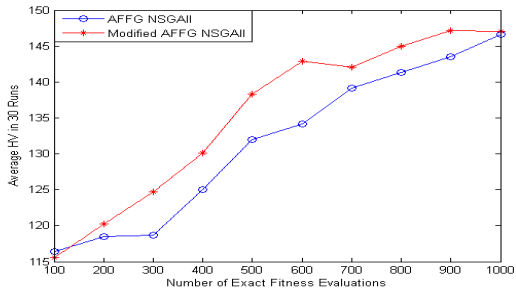

Fig.10  Performance (HV) Comparison of Modified_AFFG_NSGA2 and AFFG_NSGA2 over ZDT4 Problem.

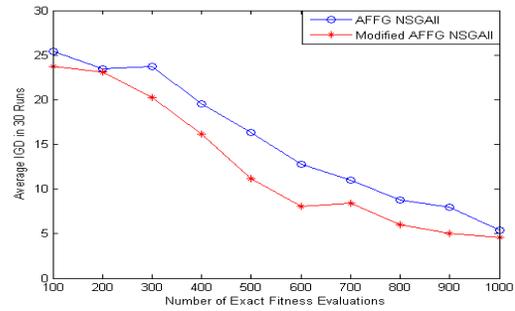

Fig.11 Performance (IGD) Comparison of Modified_AFFG_NSGA2 and AFFG_NSGA2 over ZDT4 Problem.

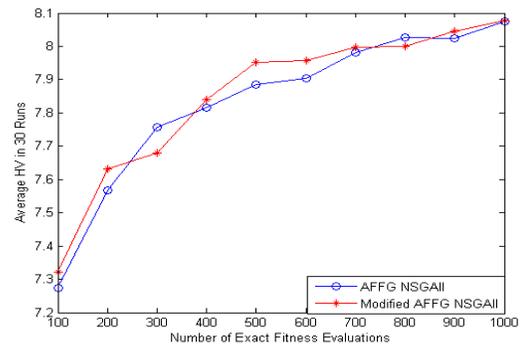

Fig.12   Performance (HV) Comparison of Modified_AFFG_NSGA2 and AFFG_NSGA2 over CF1 Problem.

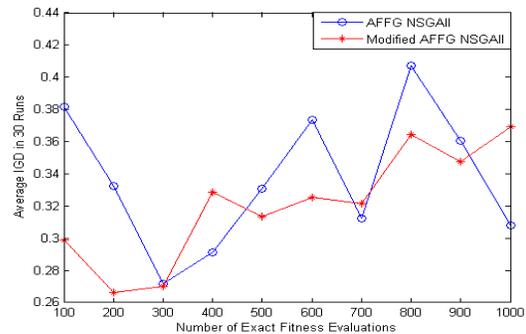

Fig.13  Performance (IGD) Comparison of Modified_AFFG_NSGA2 and AFFG_NSGA2 over CF1 Problem.

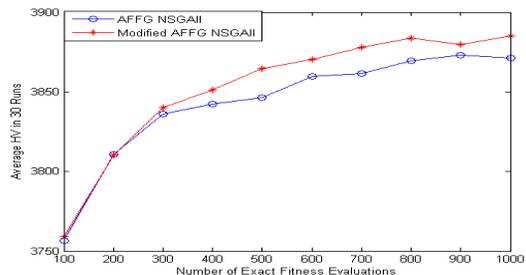

Fig.14   Performance (HV) Comparison of Modified_AFFG_NSGA2 and AFFG_NSGA2 over CF3 Problem.

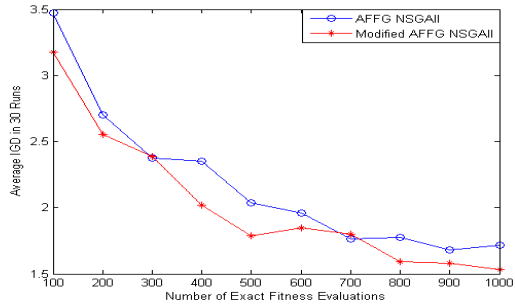

Fig.15  Performance (IGD) Comparison of Modified_AFFG_NSGA2 and AFFG_NSGA2 over CF3 Problem.

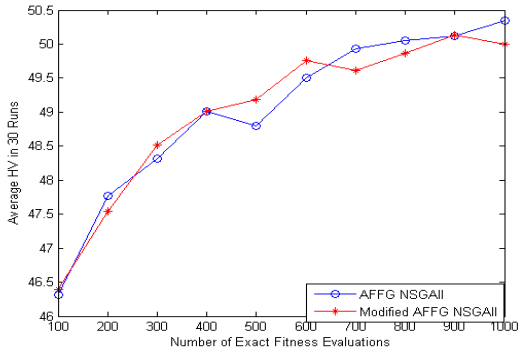

Fig.16  Performance (HV) Comparison of Modified_AFFG_NSGA2 and AFFG_NSGA2 over UF1 Problem.

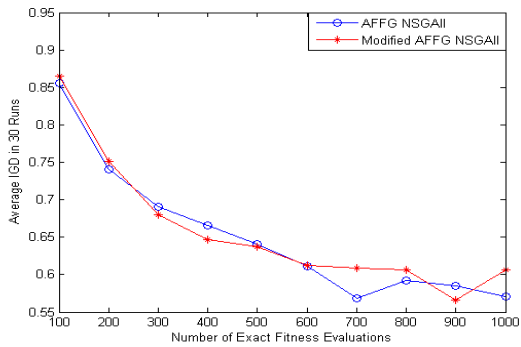

Fig.17  Performance (IGD) Comparison of Modified_AFFG_NSGA2 and AFFG_NSGA2 over UF1 Problem.

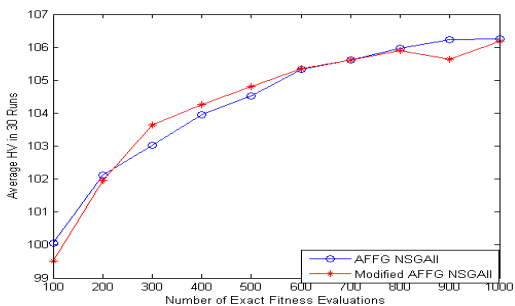

Fig.18  Performance (HV) Comparison of Modified_AFFG_NSGA2 and AFFG_NSGA2 over UF3 Problem.

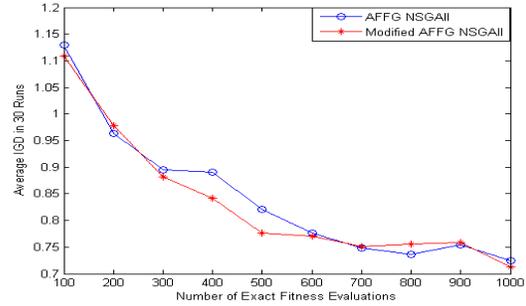

Fig.19  Performance (IGD) Comparison of Modified_AFFG_NSGA2 and AFFG_NSGA2 over UF3 Problem.

As mentioned in Section 4, both methods were run until a fixed number of generations characterized in terms of the convergence time per test problem. To have a deep analysis of the proposed approach, the average HVs, and the average IGDs (of 30 independent runs), like the average number of exact fitness function evaluations in the previous section, were plotted separately against the determined number of generations. Derivative Figures in Section 4 (Fig. 1 to Fig. 7) and those are demonstrated in this section (Fig. 20 to Fig. 31) indicate that in our proposed approach the computational cost remarkably decreases while the convergence speed reduces. Fortunately, reduction in the convergence speed is negligible in comparison with the amount of decreasing the computational cost per test problem.

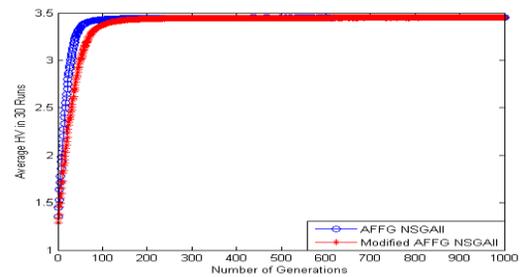

Fig.20  Performance (HV) Comparison of Modified_AFFG_NSGA2 and AFFG_NSGA2 over ZDT1 Problem.

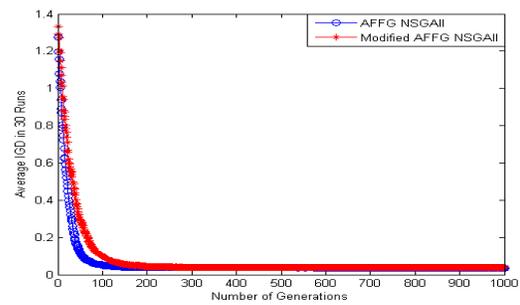

Fig.21 Performance (IGD) Comparison of Modified_AFFG_NSGA2 and AFFG_NSGA2 over ZDT1 Problem.

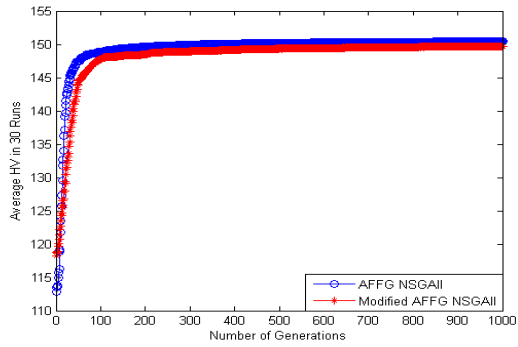

Fig.22 Performance (HV) Comparison of Modified_AFFG_NSGA2 and AFFG_NSGA2 over ZDT4 Problem.

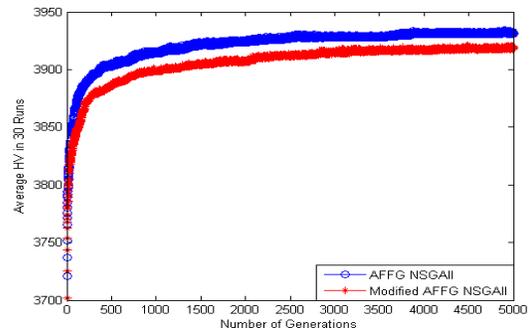

Fig.26 Performance (HV) Comparison of Modified_AFFG_NSGA2 and AFFG_NSGA2 over CF3 Problem.

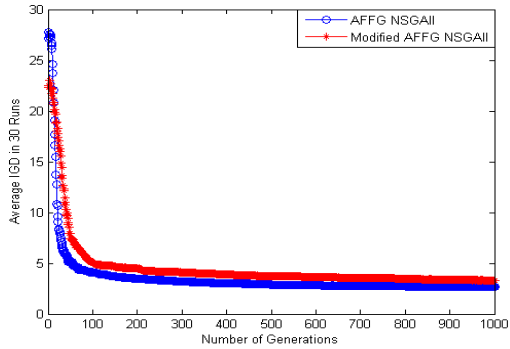

Fig.23 Performance (IGD) Comparison of Modified_AFFG_NSGA2 and AFFG_NSGA2 over ZDT4 Problem.

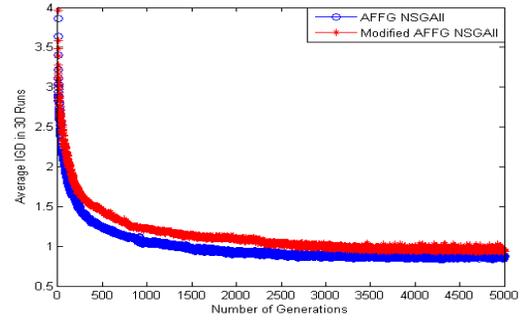

Fig.27 Performance (IGD) Comparison of Modified_AFFG_NSGA2 and AFFG_NSGA2 over CF3 Problem.

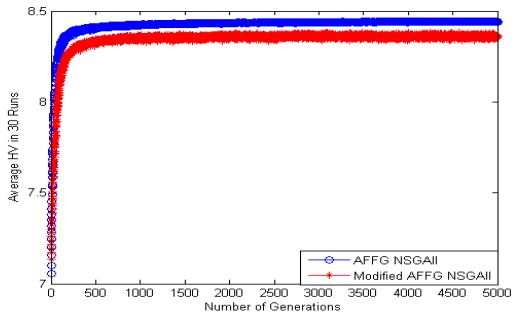

Fig.24 Performance (HV) Comparison of Modified_AFFG_NSGA2 and AFFG_NSGA2 over CF1 Problem.

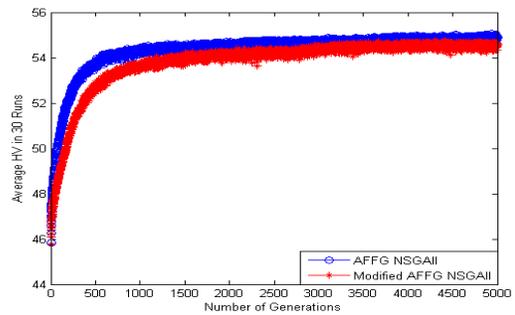

Fig.28 Performance (HV) Comparison of Modified_AFFG_NSGA2 and AFFG_NSGA2 over UF1 Problem.

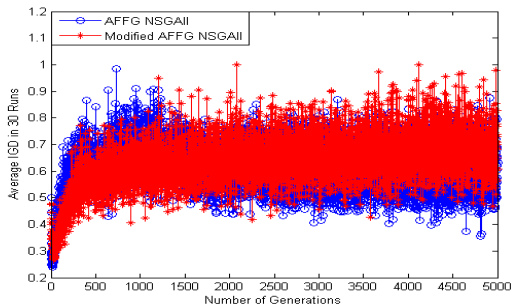

Fig.25 Performance (IGD) Comparison of Modified_AFFG_NSGA2 and AFFG_NSGA2 over CF1 Problem.

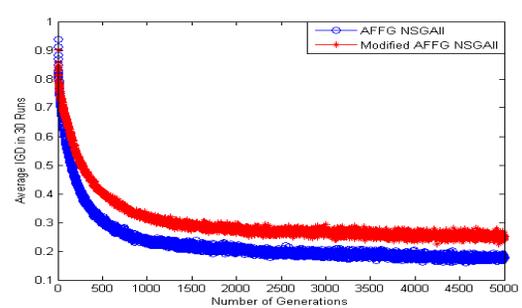

Fig.29 Performance (IGD) Comparison of Modified_AFFG_NSGA2 and AFFG_NSGA2 over UF1 Problem.

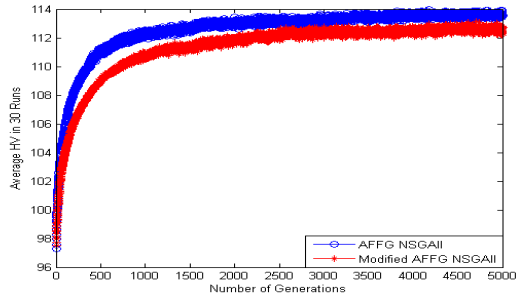

Fig.30 Performance (HV) Comparison of Modified_AFFG_NSGA2 and AFFG_NSGA2 over UF3 Problem.

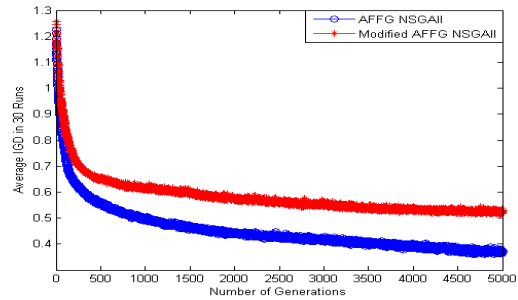

Fig.31 Performance (IGD) Comparison of Modified_AFFG_NSGA2 and AFFG_NSGA2 over UF3 Problem.

To find a better understanding of the usefulness of the proposed approach, numerical results are provided. In Tables 10, 11, and 12 the comparative amounts of the area under both AFFG_NSGAII and the Modified_AFFG_NSGAII curves (correspond to the Figures 44 to 71) are presented.

Table 10: The area under both AFFG_NSGAII and Modified_AFFG_NSGAII curves in ZDT family correspond to Figs 20 to 23.

| Problem | Average IGD – Number of Fitness Evaluations AFFG - Modified AFFG | Average HV – Number of Fitness Evaluations AFFG - Modified AFFG |
|---|---|---|
| ZDT1 | 331.1715 - 260.0680 | 2587.1 – 2726.9 |
| ZDT2 | 701.3847 -  557.1235 | 3386.3 – 3582.6 |
| ZDT3 | 563.4949 – 481.2454 | 4834.4 – 4998 |
| ZDT4 | 13893 – 11228 | 118390 – 122180 |
| ZDT6 | 5802 – 5535 | 1375.8 - 1600.4 |

Table 11: The area under both AFFG_NSGAII and Modified_AFFG_NSGAII curves in CF family correspond to Figs 24 to 27.

| Problem | Average IGD – Number of Fitness Evaluations AFFG - Modified AFFG | Average HV – Number of Fitness Evaluations AFFG - Modified AFFG |
|---|---|---|
| CF1 | 302.5250 -  287.1000 | 7063.3 - 7080.3 |
| CF2 | 523.86 - 514.0750 | 45962 - 45822 |
| CF3 | 1924.6 – 1792.2 | 3461400 - 3470000 |
| CF4 | 736.4810 - 714.8476 | 291500 -  291630 |
| CF5 | 2139 – 2120.6 | 817870 - 815850 |

Table 12: The area under both AFFG_NSGAII and Modified_AFFG_NSGAII curves in UF family correspond to Figs 28 to 31.

| Problem | Average IGD – Number of Fitness Evaluations AFFG –Modified AFFG | Average HV – Number of Fitness Evaluations AFFG – New Modified AFFG |
|---|---|---|
| UF1 | 1.1369e+003 - 1.5205e+003 | 2.6903e+0052.7172e+005 - |
| UF2 | 569.7237 - 695.9495 | 1.4502e+005 - 1.4418e+005 |
| UF3 | 2.2725e+003 - 2.9205e+003 | 5.6256e+005 - 5.5642e+005 |
| UF5 | 1.2000e+004 - 1.2880e+004 | 1.3309e+006 - 1.3243e+006 |

Moreover, to be more understandable, Tables 13, 14, and 15 signify the percentage of differences of those comparative amounts. In particular, for some test problems such as CF4 the computational cost improved to more than 43% while the convergence speed reduced to less than 1%.

Table 13: The percentage of differences of the area under the AFFG_NSGAII and the Modified_AFFG_NSGAII curves in ZDT family correspond to Figs 20 to 23 and Figs 2 to 3.

| Problem | HV – Number of Generations | Evaluations – Number of Generations |
|---|---|---|
| ZDT1 | 0.9376 | 30.1030 |
| ZDT2 | 0.3101 | 36.8103 |
| ZDT3 | 0.7941 | 33.0024 |
| ZDT4 | 0.7585 | 46.0977 |
| ZDT6 | 4.6630 | 38.4838 |

Table 14: The percentage of differences of the area under the AFFG_NSGAII and the Modified_AFFG_NSGAII curves in CF family correspond to Figs 24 to 27 and Figs 4 to 5.

| Problem | HV – Number of Generations | Evaluations – Number of Generations |
|---|---|---|
| CF1 | 0.9642 | 35.1581 |
| CF2 | 0.8242 | 30.6539 |
| CF3 | 0.3839 | 39.4853 |
| CF4 | 0.3092 | 43.6179 |
| CF6 | 0.4056 | 26.4578 |

Table 15: The percentage of differences of the area under the AFFG_NSGAII and the Modified_AFFG_NSGAII curves in UF family correspond to Figs 28 to 31 and Figs 6 to 7.

| Problem | HV – Number of Generations | Evaluations – Number of Generations |
|---|---|---|
| UF1 | 0.9906 | 39.2636 |
| UF2 | 0.5766 | 41.2803 |
| UF3 | 1.0910 | 43.0236 |
| UF4 | 0.4973 | 24.3047 |
| UF6 | 0.9906 | 39.2636 |

As it was showed in Tables 1, 2, and 3 in Section 4, the number of decision variables is considered 6, 10, and 30 for "ZDT1 to ZDT3", "ZDT4, ZDT5, and CF1 to CF5", and "UF1 to UF3 and UF5", respectively to have further investigation. The results, from first to end, illustrated that increasing individuals dimension has greater negative impact on Modified_AFFG_NSGAII form the viewpoint of both efficiency and efficacy. To explore the reason, we found the following observations. First of all, we suppose that the maximum similarity of a new offspring to the pool be more than 0.9 (predefined threshold); if the number of decision variables be set 100 in one time and 10 in the other time, the ratio of dissimilarity (between the new offspring and the granule whose similarity to it is more than predefined threshold) of the first case to the second one is 10 to 1. So, fitness is approximated with a lower accuracy in the first case. Furthermore, according to what was explained before, a greater number of fitness approximations are performed by Modefied_AFFG_NSGAII rather than AFFG_NSGAII until evolution control criterion is met. Therefore, decreasing both speed and accuracy is more tangible in our proposed approach rather than AFFG_NSGAII while the number of decision variables is increased.

## 7. Conclusion and Future Directions

In this study, we have introduced an effective factor for fitness approximation inspired from information granulation that affirmatively impress on reducing cost of MOEAs optimization. Our comprehensive experiments illustrate that the proposed approach is promising.

As a future work, we can explore some extra factors to find valuable individuals more and more precisely. Also, our proposed approach can be employed in many objective problems.

## References

[1] K. Deb, "Multi-objective Optimization Using Evolutionary Algorithms: An Introduction," Technical Report 2011003, Indian Institute of Technology Kanpur, 2011.

[2] K. Deb, "Multi-Objective Optimization Using Evolutionary Algorithms," Wiley-Interscience Series in Systems and Optimization. John Wiley & Sons, Chichester, 2001.

[3] K. Deb, "Multi-objective Evolutionary Optimization: Past, Present and Future," *Proceedings of the Fourth International Conference on Adaptive Computing in Design and Manufacture (ACDM'2000), UK, London, pp. 225–236, 2000.*

[4] G. Rohling, "Multiple Objective Evolutionary Algorithms for Independent for Independent, Computationally Expensive Objective Evaluations," PhD. Thesis, School of Electrical and Computer Engineering, Georgia Institute of Technology, Georgia(Atlanta), USA, 2004.

[5] J. E. Rodríguez, A. L. Medaglia, and C. A. Coello Coello, "Design of a Motorcycle Frame Using Neuroacceleration Strategies in MOEAs," Journal of Heuristics, Vol. 15, No. 2, 2009, pp. 177-196.

[6] M. Davarynejad, J. Rezaei, J. Vrancken, J. Berg, and C. A. Coello Coello, "Accelerating Convergence Towards the Optimal ParetoFront," Proceeding of IEEE Congress on Evolutionary Computation, , 2011, pp. 2107 – 2114.

[7] M. Davarynejad, M. R. Akbarzadeh-T, and N. Pariz, "A Novel General Framework for Evolutionary Optimization: Adaptive Fuzzy Fitness Granulation," Proceeding of IEEE Congress on Evolutionary Computing, 2007, pp. 951 – 956.

[8] Y. Jin, "A Comprehensive Survey of Fitness Approximation in Evolutionary Computation," Soft Computation, Vol. 9, No. 1, 2005, pp. 3–12.

[9] Y. Jin, "Surrogate-Assisted Evolutionary Computation: Recent Advances and Future Challenges," Swarm and Evolutionary Computation, Vol. 1, No. 2, 2011, pp. 61–70.

[10] I. Loshchilov, M. Schoenauer, and M. Sebag, "A Mono Surrogate for Multiobjective Optimization," Proceedings of the Twelfth Annual Conference on Genetic and Evolutionary Computation, 2010, pp. 471-478.

[11] I. Loshchilov, M. Schoenauer, and M. Sebag, "A Pareto-Compliant Surrogate Approach for Multiobjective Optimization," Proceedings of the Twelfth Annual Conference on Genetic and Evolutionary Computation, 2010 Vol. pp. 1979-1982.

[12] I. Loshchilov, M. Schoenauer, and M. Sebag, "Dominance-Based Pareto-Surrogate for Multi-Objective Optimization," Proceedings of the 8th international conference on Simulated evolution and learning (SEAL), USA, pp. 230-239, 2010.

[13] C.W. Seah, Y. S. Ong, I. W. Tsang, S. Jiang, "Pareto Rank Learning in Multi-objective Evolutionary Algorithms," In Proceedings of the IEEE Congress on Evolutionary Computation, 2012, pp. 1-8.

[14] M. Davarynejad, "Adaptive Fuzzy Fitness Granulation in Evolutionary Algorithms for Complex Optimization," M.S. Thesis, Department Electrical Engineering-Control Program, Ferdowsi University of Mashhad, Iran, 2007.

[15] R.E. Smith, B. A. Dike, and S. A. Stegmann, "Fitness Inheritance in Genetic Algorithms," Proceedings of the ACM symposium on Applied computing (SAC), 1995, PP. 345-350.

[16] M. Davarynejad, M. R. Akbarzadeh-T, and C. A. Coello Coello, "Auto-Tuning Fuzzy Granulation for Evolutionary Optimization," Proceedings of IEEE International Conference on Evolutionary Computation (CEC), IEEE Service Center, 2007, pp. 951–956.

[17] M. Davarynejad, C. W. Ahn, J. Vrancken, J. van den Berg, and C.A. Coello Coello, "Evolutionary hidden information detection by granulation-based fitness approximation," Soft Computing, Vol. 10, No. 3, 2010, pp. 719-729.

[18] K. Deb, A. Pratap, S. Agarwal, and T. Meyarivan, "A Fast and Elitist Multiobjective Genetic Algorithm: NSGA-II," IEEE Transactions on Evolutionary Computation, Vol. 6, No. 2, 2002, pp. 182–197.

[19] O. Schuetze, X. Esquivel, A. Lara, and C. A. Coello Coello, "Some Comments on GD and IGD and Relations to the Hausdorff Distance n," In proceeding of Genetic and Evolutionary Computation Conference, 2010, pp.1971-1974.

[20] Q. Zhang, A. Zhou, S. Zhao, P. N. Suganthany, W. Liu, and S. Tiwari, "Multiobjective optimization Test Instances for the CEC 2009 Special Session and Competition," Proceeding

of IEEE Congress on Evolutionary Computing, 2009, pp.18-21.
[21] E. Zitzler, K. Deb, and L. Thiele, "Comparison of Multiobjective Evolutionary Algorithms: Empirical Results," Evolutionary Computation, Vol. 8, No. 2, 2000, pp. 173–195.

**Z. Pourbahman** is the MSc student in artificial intelligence of CSE and IT Department of Shiraz University. As one of her research interests, she focuses on evolutionary algorithms specially fitness approximation domain. Also, she is co-author of several articles in Machine Translation area especially in Natural Language Processing domain. She was programmer team leader in Telecommunication Company of Iran in 2008.

**A. Hamzeh** received his Ph.D. in artificial intelligence from Iran University of Science and Technology (IUST) in 2007. Since then, he has been working as assistant professor in CSE and IT Department of Shiraz University. There, he is one of the founders of local CERT center which serves as the security and protection service provider in its local area. As one of his research interests, he recently focuses on cryptography and steganography area and works as a team leader in CERT center to develop and break steganography method, especially in image spatial domain. Also, he works as one of team leaders of Soft Computing group of shiraz university working on bio-inspired optimization algorithms. He is co-author of several articles in security and optimization.